# Reconfiguring the Imaging Pipeline for Computer Vision

Mark Buckler
Cornell University

Suren Jayasuriya
Carnegie Mellon University

Adrian Sampson
Cornell University

## Abstract

*Advancements in deep learning have ignited an explosion of research on efficient hardware for embedded computer vision. Hardware vision acceleration, however, does not address the cost of capturing and processing the image data that feeds these algorithms. We examine the role of the image signal processing (ISP) pipeline in computer vision to identify opportunities to reduce computation and save energy. The key insight is that imaging pipelines should be be configurable: to switch between a traditional* photography mode *and a low-power* vision mode *that produces lower-quality image data suitable only for computer vision. We use eight computer vision algorithms and a reversible pipeline simulation tool to study the imaging system's impact on vision performance. For both CNN-based and classical vision algorithms, we observe that only two ISP stages, demosaicing and gamma compression, are critical for task performance. We propose a new image sensor design that can compensate for these stages. The sensor design features an adjustable resolution and tunable analog-to-digital converters (ADCs). Our proposed imaging system's vision mode disables the ISP entirely and configures the sensor to produce subsampled, lower-precision image data. This vision mode can save ~75% of the average energy of a baseline photography mode with only a small impact on vision task accuracy.*

## 1. Introduction

The deep learning revolution has accelerated progress in a plethora of computer vision tasks. To bring these vision capabilities within the battery budget of a smartphone, a wave of recent work has designed custom hardware for inference in deep neural networks [16, 21, 34]. This work, however, only addresses part of the whole cost: embedded vision involves the entire imaging pipeline, from photons to task result. As hardware acceleration reduces the energy cost of inference, the cost to capture and process images will consume a larger share of total system power [9, 35].

We study the potential for co-design between camera systems and vision algorithms to improve their end-to-end efficiency. Existing imaging pipelines are designed for photography: they produce high-quality images for human consumption. An imaging pipeline consists of the image sensor itself and an *image signal processor* (ISP) chip, both of which are hard-wired to produce high-resolution, low-noise, color-corrected photographs. Modern computer vision algorithms, however, do not require the same level of quality that humans do. Our key observation is that mainstream, photography-oriented imaging hardware wastes time and energy to provide quality that computer vision algorithms do not need.

We propose to make imaging pipelines configurable. The pipeline should support both a traditional *photography mode* and an additional, low-power *vision mode*. In vision mode, the sensor can save energy by producing lower-resolution, lower-precision image data, and the ISP can skip stages or disable itself altogether. We examine the potential for a vision mode in imaging systems by measuring its impact on the hardware efficiency and vision accuracy. We study vision algorithms' sensitivity to sensor parameters and to individual ISP stages, and we use the results to propose an end-to-end design for an imaging pipeline's vision mode.

**Contributions:** This paper proposes a set of modifications to a traditional camera sensor to support a vision mode. The design uses variable-accuracy analog-to-digital converters (ADCs) to reduce the cost of pixel capture and power-gated selective readout to adjust sensor resolution. The sensor's subsampling and quantization hardware approximates the effects of two traditional ISP stages, demosaicing and gamma compression. With this augmented sensor, we propose to disable the ISP altogether in vision mode.

We also describe a methodology for studying the imaging system's role in computer vision performance. We have developed a tool that simulates a configurable imaging pipeline and its inverse to convert plain images to approximate raw signals. This tool is critical for generating training data for learning-based vision algorithms that need examples of images produced by a hypothetical imaging pipeline. Section 3.2 describes the open-source simulation infrastructure.

We use our methodology to examine eight vision applications, including classical algorithms for stereo, optical flow, and structure-from-motion; and convolutional neural networks (CNNs) for object recognition and detection. For these applications, we find that:

- Most traditional ISP stages are unnecessary when targeting computer vision. For all but one application we tested, only two stages had significant effects on vision accuracy: demosaicing and gamma compression.
- Our image sensor can approximate the effects of demosaicing and gamma compression in the mixed-signal domain. Using these in-sensor techniques eliminates the need for a separate ISP for most vision applications.
- Our image sensor can reduce its bitwidth from 12 to 5 by replacing linear ADC quantization with logarithmic quantization while maintaining the same level of task performance.

Altogether, the proposed vision mode can use roughly a quarter of the imaging-pipeline energy of a traditional photography mode without significantly affecting the performance of most vision algorithms we studied.

## 2. Related Work

**Energy-efficient Deep Learning:** Recent research has focused on dedicated ASICs for deep learning [10, 16, 21, 34, 43**?** ] to reduce the cost of forward inference compared to a GPU or CPU. Our work complements this agenda by focusing on energy efficiency in the rest of the system: we propose to pair low-power vision implementations with low-power sensing circuitry.

**ISPs for Vision:** While most ISPs are fixed-function designs, Vasilyev et al. [50] propose to use a programmable CGRA architecture to make them more flexible, and other work has synthesized custom ISPs onto FPGAs [24, 25]. Mainstream cameras, including smartphones [2], can bypass the ISP to produce RAW images, but the associated impact on vision is not known. Liu et al. [36] propose an ISP that selectively disables stages depending on application needs. We also explore sensitivity to ISP stages, and we propose changes to the image sensor hardware that subsume critical stages in a traditional ISP.

**Image Sensors for Vision:** In industry, some cameras are marketed with vision-specific designs. For example, Centeye [5] offers image sensors based on a logarithmic-response pixel circuit [17] for high dynamic range. Omid-Zohoor et al. [39] propose logarithmic, low-bitwidth ADCs and on-sensor processing for efficient featurization using the histogram of oriented gradients. *Focal-plane processing* can compute basic functions such as edge detection in analog on the sensor [11, 37]. Red-Eye [34] computes initial convolutions for a CNN using a custom sensor ADC, and Chen et al. [8] approximate the first layer *optically* using angle-sensitive pixels. Event-based vision sensors detect temporal motion with custom pixels [3, 27]. Chakrabarti [7] proposes to learn novel, non-Bayer sensor layouts using backpropagation. We focus instead on minimally invasive changes to existing camera pipelines. To our knowledge, this is the first work to measure vision applications' sensitivity to design decisions in a traditional ISP pipeline. Our proposed pipeline can support both computer vision and traditional photography.

Other work has measured the energy of image sensing: there are potential energy savings when adjusting a sensor's frame rate and resolution [35]. Lower-powered image sensors have been used to decide when to activate traditional cameras and full vision computations [22].

Compressive sensing shares our goal of reducing sensing cost, but it relies on complex computations to recover images [15]. In contrast, our proposed pipeline lets vision algorithms work directly on sensor data without additional image reconstruction.

**Error Tolerance in CNNs:** Recent work by Diamond et al. [14] studies the impact of sensor noise and blurring on CNN accuracy and develops strategies to tolerate it. Our focus is broader: we consider a range of sensor and ISP stages, and we measure both CNN-based and "classical" computer vision algorithms.

## 3. Background & Experimental Setup

### 3.1. The Imaging Pipeline

Figure 1a depicts a traditional imaging pipeline that feeds a vision application. The main components are an image sensor, which reacts to light and produces a RAW image; an image signal processor (ISP) unit, which transforms, enhances, and compresses the signal to produce a complete image, usually in JPEG format; and the vision application itself.

ISPs consist of a series of signal processing stages. While the precise makeup of an ISP pipeline varies, we consider a typical set of stages common to all ISP pipelines: denoising, demosaicing, color transformations, gamut mapping, tone mapping, and image compression. This simple pipeline is idealized: modern ISPs can comprise hundreds of proprietary stages. For example, tone mapping and denoising can use complex, adaptive operations that are customized for specific camera hardware. In this paper, we consider a simple form of global tone mapping that performs *gamma com-*

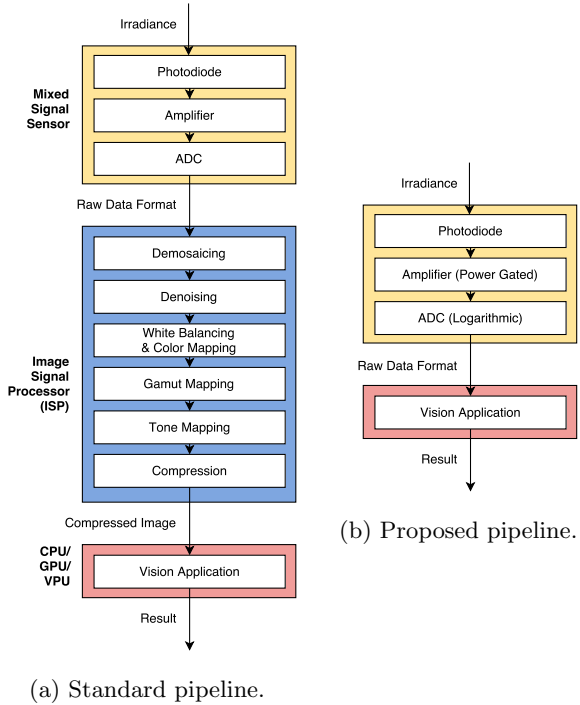

(a) Standard pipeline.

(b) Proposed pipeline.

Figure 1: The standard imaging pipeline (a) and our proposed pipeline (b) for our design's *vision mode*.

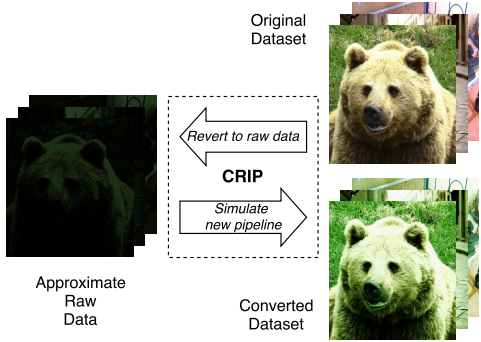

Figure 2: Configurable & Reversible Imaging Pipeline.

*pression*. We also omit analyses that control the sensor, such as autoexposure and autofocus, and specialized stages such as burst photography or high dynamic range (HDR) modes. We select these simple, essential ISP stages because we believe they represent the common functionality that may impact computer vision.

### 3.2. Pipeline Simulation Tool

Many computer vision algorithms rely on machine learning. Deep learning techniques in particular require vast bodies of training images. To make learning-based vision work on our proposed imaging pipelines, we need a way to generate labeled images that look as if they were captured by the hypothetical hardware. Instead of capturing this data from scratch, we develop a toolchain that can *convert* existing image datasets.

The tool, called the Configurable & Reversible Imaging Pipeline (CRIP), simulates an imaging pipeline in "forward" operation and inverts the function in "reverse" mode. CRIP takes as input a standard image file, runs the inverse conversion to approximate a RAW image, and then simulates a specific sensor/ISP configuration to produce a final RGB image. The result recreates the image's color, resolution and quantization as if it had been captured and processed by a specific image sensor and ISP design. Figure 2 depicts the workflow and shows the result of simulating a pipeline with only gamma compression and demosaicing. Skipping color transformations leads to a green hue in the output.

The inverse conversion uses an implementation of Kim et al.'s reversible ISP model [31] augmented with new stages for reverse denoising and demosaicing as well as re-quantization. To restore noise to a denoised image, we use Chehdi et al.'s sensor noise model [48]. To reverse the demosaicing process, we remove channel data from the image according to the Bayer filter. The resulting RAW image approximates the unprocessed output of a camera sensor, but some aspects cannot be reversed: namely, sensors typically digitize 12 bits per pixel, but ordinary 8-bit images have lost this detail after compression. For this reason, we only report results for quantization levels with 8 bits or fewer.

CRIP implements the reverse stages from Kim et al. [31], so its model linearization error is the same as in that work: namely, less than 1%. To quantify CRIP's error when reconstructing RAW images, we used it to convert a Macbeth color chart photograph and compared the result with its original RAW version. The average pixel error was 1.064% and the PSNR was 28.81 dB. Qualitatively, our tool produces simulated RAW images that are visually indistinguishable from their real RAW counterparts.

CRIP's reverse pipeline implementation can use any camera model specified by Kim et al. [31], but for consistency, this paper uses the Nikon D7000 pipeline. We have implemented the entire tool in the domain-specific language Halide [42] for speed. For example, CRIP can convert the entire CIFAR-10 dataset [32] in one hour on an 8-core machine. CRIP is available as open source: https://github.com/cucapra/approx-vision

### 3.3. Benchmarks

Table 1 lists the computer vision algorithms we study. It also shows the data sets used for evaluation and, where applicable, training. Our suite consists of 5 CNN-based algorithms and 3 "classical," non-learning implementations covering a range of vision tasks: object classification, object detection, face identification, optical flow, and structure from motion. For object classification, we test 3 different implementations of varying sophistication to examine the impact of neural network depth on error tolerance.

| Algorithm | Dataset | Vision Task |
|---|---|---|
| 3 Deep LeNet [33] | CIFAR-10 [32] | Obj. Classification |
| 20 Deep ResNet [23] | CIFAR-10 | Obj. Classification |
| 44 Deep ResNet [23] | CIFAR-10 | Obj. Classification |
| Faster R-CNN [44] | VOC-2007 [19] | Object Detection |
| OpenFace [1] | CASIA [51] and LFW [26] | Face Identification |
| OpenCV Farneback [28] | Middlebury [46] | Optical Flow |
| OpenCV SGBM [28] | Middlebury | Stereo Matching |
| OpenMVG SfM [38] | Strecha [47] | Structure from Motion |

Table 1: Vision applications used in our evaluation.

For each experiment, we configure CRIP to apply a chosen set of ISP stages and to simulate a given sensor resolution and ADC quantization. For the CNNs, we convert a training set and train the network starting with pre-trained weights using the same learning rates and hyperparameters specified in the original paper. For all applications, we convert a test set and evaluate performance using an algorithm-specific metric.

## 4. Sensitivity to ISP Stages

We next present an empirical analysis of our benchmark suite's sensitivity to stages in the ISP. The goal is to measure, for each algorithm, the relative difference in task performance between running on the original image data and running on data converted by CRIP.

**Individual Stages:** First, we examine the sensitivity to each ISP stage in isolation. Testing the exponential space of all possible stage combinations is intractable, so we start with two sets of experiments: one that *disables* a single ISP stage and leaves the rest of the pipeline intact (Figure 3a); and one that *enables* a single ISP stage and disables the rest (Figure 3b).

In these experiments, gamut mapping and color transformations have a minimal effect on all benchmarks. The largest effects are on ResNet44, where classification error increases from 6.3% in the baseline to 6.6% without gamut mapping, and OpenMVG, where removing the color transform stage increases RMSE from 0.408 to 0.445. This finding confirms that features for vision are not highly sensitive to color.

There is a strong sensitivity, in contrast, to gamma compression and demosaicing. The OpenMVG Structure from Motion (SfM) implementation fails entirely when gamma compression is disabled: it was unable to find sufficient features using either of its feature extractors, SIFT and AKAZE. Meanwhile, removing demosaicing worsens the error for Farneback optical flow by nearly half, from 0.227 to 0.448. Both of these classical (non-CNN) algorithms use hand-tuned feature extractors, which do not take the Bayer pattern into account. The CIFAR-10 benchmarks (LeNet3, ResNet20, ResNet44) use low-resolution data (32×32), which is disproportionately affected by the removal of color channels in mosaiced data. While gamma-compressed data follows a normal distribution, removing gamma compression reverts the intensity scale to its natural log-normal distribution, which makes features more difficult to detect for both classical algorithms and CNNs.

Unlike the other applications, Stereo SGBM is sensitive to noise. Adding sensor noise increases its mean error from 0.245 to 0.425, an increase of over 70%. Also unlike other applications, OpenFace counter-intuitively performs *better* than the baseline when the simulated pipeline omits gamut mapping or gamma compression. OpenFace's error is 8.65% on the original data and 7.9% and 8.13%, respectively, when skipping those stages. We attribute the difference to randomness inherent in the training process. Across 10 training runs, OpenFace's baseline error rate varied from 8.22% to 10.35% with a standard deviation of 0.57%.

**Minimal Pipelines:** Based on these results, we study the effect of combining the most important stages: demosaicing, gamma compression, and denoising. Figure 4 shows two configurations that enable only the first two and all three of these stages. Accuracy for the minimal pipeline with only demosaicing and gamma compression is similar to accuracy on the original data. The largest impact, excluding SGBM, is ResNet44, whose top-1 error increases only from 6.3% to 7.2%. Stereo SGBM, however, is noise sensitive: without denoising, its mean error is 0.33; with denoising, its error returns to its baseline of 0.25. Overall, the CNNs are able to rely on retraining themselves to adapt to changes in the capture pipeline, while classical benchmarks are less flexible and can depend on specific ISP stages.

We conclude that demosaicing and gamma compression are the only important stages for all applications except for one, which also benefits from denoising. Our goal in the next section is to show how to remove the need for these two stages to allow vision mode to disable the ISP entirely. For outliers like SGBM, selectively enabling the ISP may still be worthwhile.

## 5. Image Sensor Design

Based on our experiments with limited ISP processing, we propose a new image sensor design that can

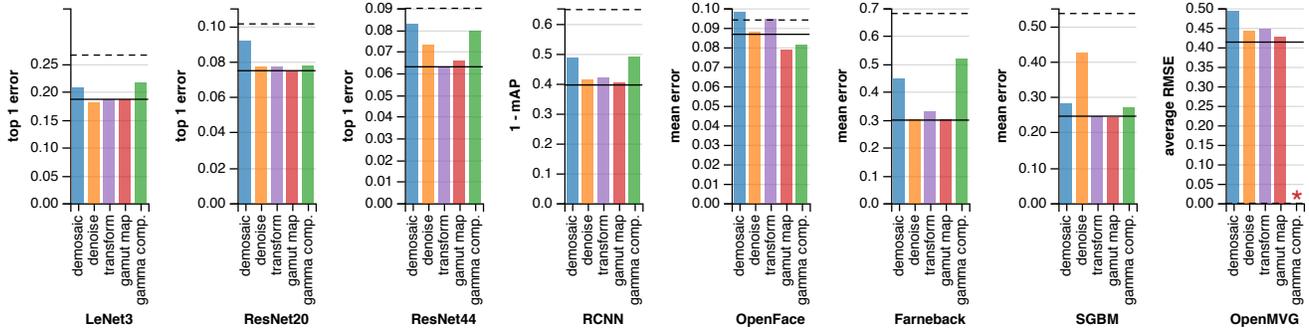

(a) Disabling a single ISP stage.

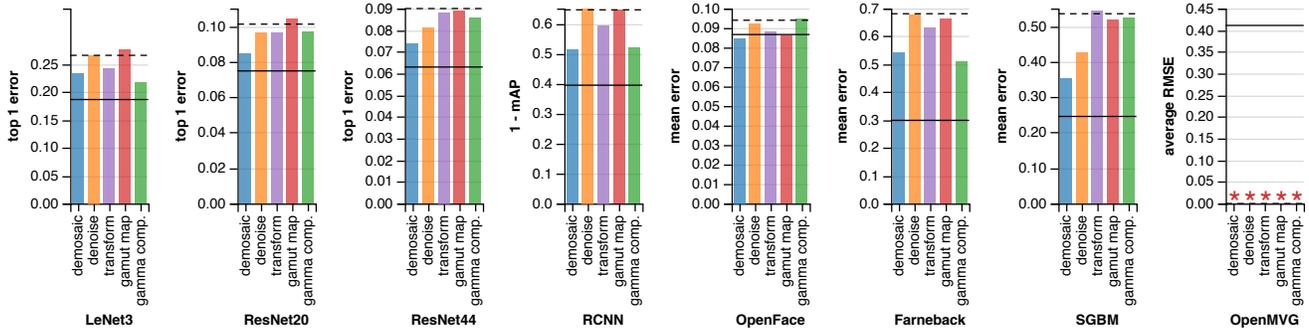

(b) Enabling a single ISP stage and disabling the rest.

Figure 3: The impact on vision accuracy when adding and removing stages from the standard ISP pipeline. The solid line shows the baseline error with all ISP stages enabled, and the dotted line shows the error when all ISP stages are disabled. Asterisks denote aborted runs where no useful output was produced.

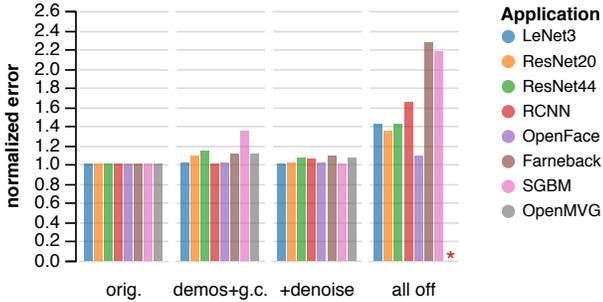

Figure 4: Each algorithm's vision error, normalized to the original error on plain images, for two minimal ISP pipelines. The *demos+g.c.* pipeline only enables demosaicing and gamma compression; the *+denoise* bars also add denoising. The *all off* column shows a configuration with all stages disabled for reference.

operate in a low-power vision mode. We propose three key features: adjustable resolution via selective pixel readout and power gating; subsampling to approximate ISP-based demosaicing; and nonlinear ADC quantization to perform gamma compression. All three are well-known sensor design techniques; we propose to use them in an optional camera mode to replace the ISP's role in embedded vision applications.

**Resolution:** A primary factor in a sensor's energy consumption is the resolution. Frames are typically read out in column-parallel fashion, where each column of pixels passes through amplification and an ADC. Our design can selectively read out a region of interest (ROI) or subset of pixels, and save energy, by power-gating column amplifiers and ADCs. Figure 5a depicts the power-gating circuitry. The image sensor's controller unit can turn the additional transistor on or off to control power for the amplifier and ADC in each column.

**Subsampling:** Section 4 finds that most vision tasks depend on demosaicing for good accuracy. There are many possible demosaicing techniques, but they are typically costly algorithms optimized for perceived image quality. We hypothesize that, for vision algorithms, the nuances of advanced demosaicing techniques are less important than the image format: raw images exhibit the Bayer pattern, while demosaiced images use a standard RGB format.

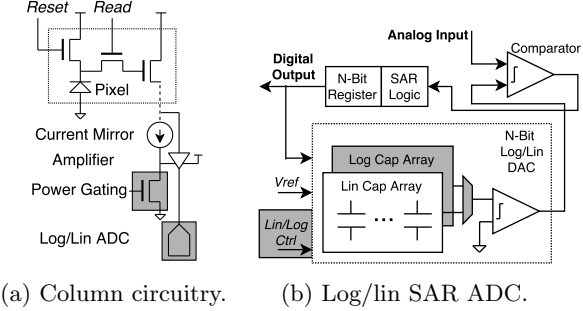

(a) Column circuitry.   (b) Log/lin SAR ADC.

Figure 5: Our proposed camera sensor circuitry, including power gating at the column level (a) and our configurable logarithmic/linear SAR ADC (b).

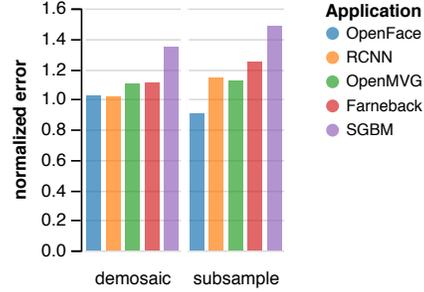

Figure 6: Demosaicing on the ISP vs. subsampling in the sensor. Error values are normalized to performance on unmodified image data.

We propose to modify the image sensor to achieve the same format-change effect as demosaicing without any signal processing. Specifically, our camera's vision mode *subsamples* the raw image to collapse each 2×2 block of Bayer-pattern pixels into a single RGB pixel. Each such block contains one red pixel, two green pixels, and one blue pixel; our technique drops one green pixel and combines it with the remaining values to form the three output channels. The design power-gates one of the two green pixels interprets resulting red, green, and blue values as a single pixel.

**Nonlinear Quantization:** In each sensor column, an analog-to-digital (ADC) converter is responsible for quantizing the analog output of the amplifier to a digital representation. A typical linear ADC's energy cost is exponential in the number of bits in its output: an 12-bit ADC costs roughly twice as much energy as a 11-bit ADC. There is an opportunity to drastically reduce the cost of image capture by reducing the number of bits.

As with resolution, ADC quantization is typically fixed at design time. We propose to make the number of bits configurable for a given imaging mode. Our proposed image sensor uses *successive-approximation* (SAR) ADCs, which support a variable bit depth controlled by a clock and control signal [49].

ADC design can also provide a second opportunity: to change the *distribution* of quantization levels. Nonlinear quantization can be better for representing images because their light intensities are not uniformly distributed: the probability distribution function for intensities in natural images is log-normal [45]. To preserve more information about the analog signal, SAR ADCs can use quantization levels that map the intensities uniformly among *digital* values. (See the supplementary material for a more complete discussion of intensity distributions.) We propose an ADC that uses logarithmic quantization in vision mode. Figure 5b shows the ADC design, which can switch between linear quantization levels for photography mode and logarithmic quantization for vision mode. The design uses a separate capacitor bank for each quantization scheme.

Logarithmic quantization lets the camera capture the same amount of image information using fewer bits, which is the same goal usually accomplished by the gamma compression stage in the ISP. Therefore, we eliminate the need for a separate ISP block to perform gamma compression.

**System Considerations:** Our proposed vision mode controls three sensor parameters: it enables subsampling to produce RGB images; it allows reduced-resolution readout; and it enables a lower-precision logarithmic ADC configuration. The data is sent off-chip directly to the application on the CPU, the GPU, or dedicated vision hardware without being compressed. This mode assumes that the vision task is running in real time, so the image does not need to be saved.

In the traditional photography mode, we configure the ADCs to be at high precision with linear quantization levels. Then the image is sent to the separate ISP chip to perform all the processing needed to generate high quality images. These images are compressed using the JPEG codec on-chip and stored in memory for access by the application processor.

## 6. Sensitivity to Sensor Parameters

We empirically measure the vision performance impact of the design decisions in our camera's vision mode. We again use the CRIP tool to simulate specific sensor configurations by converting image datasets and evaluate the effects on the benchmarks in Table 1.

**Approximate Demosaicing with Subsampling:** We first study subsampling as an alternative to true demosaicing in the ISP. In this study, we omit the benchmarks that work on CIFAR-10 images [32] because their resolution, 32×32, is unrealistically small for a sensor, so subsampling beyond this size is not meaningful. Fig-ure 10 compares data for "true" demosaicing, where CRIP has not reversed that stage, to a version that simulates our subsampling instead. Replacing demosaicing with subsampling leads to a small increase in vision error. Farneback optical flow sees the largest error increase, from 0.332 to 0.375.

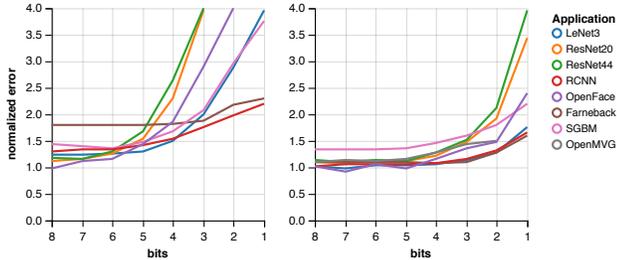

(a) Linear quantization.   (b) Logarithmic quantization.

Figure 7: Effect of quantization on vision accuracy in a pipeline with only demosaicing enabled.

**Quantization:** Next, we study the impact of signal quantization in the sensor's ADCs. There are two parameters: the number of bits and the level distribution (linear or logarithmic). Figure 13 shows our vision applications' sensitivity to both bitwidth and distribution. Both sweeps use an ISP pipeline with demosiacing but without gamma compression to demonstrate that the logarithmic ADC, like gamma compression, compresses the data distribution.

The logarithmic ADC yields higher accuracy on all benchmarks than the linear ADC with the same bitwidth. Farneback optical flow's sensitivity is particularly dramatic: using a linear ADC, its mean error is 0.54 and 0.65 for 8 and 2 bits, respectively; while with a logarithmic ADC, the error drops to 0.33 and 0.38.

Switching to a logarithmic ADC also increases the applications' tolerance to smaller bitwidths. All applications exhibit minimal error increases down to 5 bits, and some can even tolerate 4- or 3-bit quantization. OpenMVG's average RMSE only increases from 0.454 to 0.474 when reducing 8 bit logarithmic sampling to 5 bits, and ResNet20's top-1 error increases from 8.2% to 8.42%. To fit all of these applications, we propose a 5-bit logarithmic ADC design in vision mode.

**Resolution:** We next measure the impact of resolution adjustment using column power gating. Modern image sensors use multi-megapixel resolutions, while the input dimensions for most convolutional neural networks are often 256×256 or smaller. While changing the input dimensions of the network itself may also be an option, we focus here on downsampling images to match the network's published input size.

To test the downsampling technique, we concocted a new higher-resolution dataset by selecting a subset of ImageNet [13] which contains the CIFAR-10 [32] object classes (~15,000 images). These images are higher resolution than the input resolution of networks trained on CIFAR-10, so they let us experiment with image downsampling.

We divide the new dataset into training and testing datasets using an 80–20 balance and train the LeNet, ResNet20, and ResNet44 networks from pre-trained weights. For each experiment, we first downsample the images to simulate sensor power gating. Then, after demosaicing, we scale down the images the rest of the way to 32×32 using OpenCV's edge-aware scaling [28]. Without any subsampling, LeNet achieves 39.6% error, ResNet20 26.34%, and ResNet44 24.50%. We then simulated downsampling at ratios of ¼, ¹⁄₁₆, and ¹⁄₆₄ resolution. Downsampling does increase vision error, but the effect is small: the drop in accuracy from full resolution to ¼ resolution is approximately 1% (LeNet 40.9%, ResNet20 27.71%, ResNet44 26.5%). Full results are included in this paper's supplemental material.

## 7. Quantifying Power Savings

Here we estimate the potential power efficiency benefits of our proposed vision mode as compared to a traditional photography-oriented imaging pipeline. Our analysis covers the sensor's analog-to-digital conversion, the sensor resolution, and the ISP chip.

**Image Sensor ADCs:** Roughly half of a camera sensor's power budget goes to readout, which is dominated by the cost of analog-to-digital converters (ADCs) [6]. While traditional sensors use 12-bit linear ADCs, our proposal uses a 5-bit logarithmic ADC.

To compute the expected value of the energy required for each ADC's readout, we quantify the probability and energy cost of each digital level that the ADC can detect. The expected value for a single readout is:

$$\mathrm{E}\left[\mathrm{ADC\_energy}\right] = \sum_{m=1}^{2^n} p_m e_m$$

where $n$ is the number of bits, $2^n$ is the total number of levels, $m$ is the level index, $p_m$ is the probability of level $m$ occuring, and $e_m$ is the energy cost of running the ADC at level $m$.

To find $p_m$ for each level, we measure the distribution of values from images in the CIFAR-10 dataset [32] in raw data form converted by CRIP (Section 3.2). To find a relative measure for $e_m$, we simulate the operation of the successive approximation register (SAR) ADC in charging and discharging the capacitors in its bank. This capacitor simulation is a simple first-order model of a SAR ADC's power that ignores fixed overheads such as control logic.

In our simulations, the 5-bit logarithmic ADC uses 99.95% less energy than the baseline 12-bit linear ADC. As the ADCs in an image sensor account for 50% of the energy budget [6], this means that the cheaper ADCs save approximately half of the sensor's energy cost.

**Image Sensor Resolution:** An image sensor's readout, I/O, and pixel array together make up roughly 95% of its power cost [6]. These costs are linearly related to the sensor's total resolution. As Section 5 describes, our proposed image sensor uses selective readout circuitry to power off the pixel, amplifier, and ADC for subsets of the sensor array. The lower-resolution results can be appropriate for vision algorithms that have low-resolution inputs (Section 6). Adjusting the proposed sensor's resolution parameter therefore reduces the bulk of its power linearly with the pixel count.

**ISP:** While total power consumption numbers are available for commercial and research ISP designs, we are unaware of a published breakdown of power consumed per stage. To approximate the relative cost for each stage, we measured software implementations of each using OpenCV 2.4.8 [28] and profile them when processing a 4288×2848 image on an Intel Ivy Bridge i7-3770K CPU. We report the number of dynamic instructions executed, the CPU cycle count, the number of floating-point operations, and the L1 data cache references in a table in our supplementary material.

While this software implementation does not directly reflect hardware costs in a real ISP, we can draw general conclusions about relative costs. The denoising stage is by far the most expensive, requiring more than two orders of magnitude more dynamic instructions. Denoising—here, non-local means [4]—involves irregular and non-local references to surrounding pixels. JPEG compression is also expensive; it uses a costly discrete cosine transform for each macroblock.

Section 4 finds that most stages of the ISP are unnecessary in vision mode, and Section 5 demonstrates how two remaining stages—gamma compression and demosaicing—can be approximated using in-sensor techniques. The JPEG compression stage is also unnecessary in computer vision mode: because images do not need to be stored, they do not need to be compressed for efficiency. Therefore, the pipeline can fully bypass the ISP when in vision mode. Power-gating the integrated circuit would save all of the energy needed to run it.

**Total Power Savings:** The two components of an imaging pipeline, the sensor and the ISP, have comparable total power costs. For sensors, typical power costs range from 137.1 mW for a security camera to 338.6 mW for a mobile device camera [35]. Industry ISPs can range from 130 mW to 185 mW when processing 1.2 MP at 45 fps [40], while Hegarty et al. [24] simulated an automatically synthesized ISP which consumes 250 mW when processing 1080p video at 60 fps. This power consumption is comparable to recent CNN ASICs such as TrueNorth at 204 mW [18] and EIE at 590 mW [21].

In vision mode, the proposed image sensor uses half as much energy as a traditional sensor by switching to a 5-bit logarithmic ADC. The ISP can be disabled entirely. Because the two components contribute roughly equal parts to the pipeline's power, the entire vision mode saves around 75% of a traditional pipeline's energy. If resolution can be reduced, energy savings can be higher.

This first-order energy analysis does not include overheads for power gating, additional muxing, or off-chip communication. We plan to measure complete implementations in future work.

## 8. Discussion

We advocate for adding a *vision mode* to the imaging pipelines in mobile devices. We show that design choices in the sensor's circuitry can obviate the need for an ISP when supplying a computer vision algorithm.

This paper uses an empirical approach to validate our design for a vision-mode imaging pipeline. This limits our conclusions to pertain to specific algorithms and specific datasets. Follow-on work should take a theoretical approach to model the statistical effect of each ISP stage. Future work should also complete a detailed hardware design for the proposed sensor modifications. This paper uses a first-order energy evaluation that does not quantify overheads; a full design would contend with the area costs of additional components and the need to preserve pixel pitch in the column architecture. Finally, the proposed vision mode consists of conservative changes to a traditional camera design and no changes to vision algorithms themselves. This basic framework suggests future work on deeper co-design between camera systems and computer vision algorithms. By modifying the abstraction boundary between hardware and software, co-designed systems can make sophisticated vision feasible in energy-limited mobile devices.

## 9. Acknowledgements

Many thanks to Alyosha Molnar and Christopher Batten for their feedback and to Taehoon Lee and Omar Abdelaziz for hacking. This work was supported by a 2016 Google Faculty Research Award. Suren Jayasuriya was supported by an NSF Graduate Research Fellowship and a Qualcomm Innovation Fellowship.

## A. Background: The Imaging Pipeline

For readers unfamiliar with the ISP pipeline, we describe the standard pipeline found in any modern camera, from DSLRs to smartphones. This expands on Section 3.1 in the main paper.

We consider a complete system including a computer vision algorithm that processes images and produces vision results. Figure 1a in the main paper depicts the traditional pipeline. The main components are an image sensor, which reacts to light and produces a RAW image signal; an image signal processor (ISP) unit, which transforms, enhances, and compresses the signal to produce a complete image, usually in JPEG format; and the vision application itself.

### A.1. Camera Sensor

The first step in statically capturing a scene is to convert light into an electronic form. Both CCD and CMOS image sensors use solid state devices which take advantage of the photoelectric effect to convert light into voltage. Most modern devices use CMOS sensors, which use active arrays of photodiodes to convert light to charge, and then to convert charge to voltage. These pixels are typically the size of a few microns, with modern mobile image sensors reaching sizes of 1.1 µm, and are configured in arrays consisting of several megapixels.

CMOS photodiodes have a broadband spectral response in visible light, so they can only capture monochrome intensity data by themselves. To capture color, sensors add photodiode-sized filters that allow specific wavelengths of light to pass through. Each photodiode is therefore statically allocated to sense a specific color: typically red, green, or blue. The layout of these filters is called the *mosaic*. The most common mosaic is the Bayer filter [41], which is a 2×2 pattern consisting of two green pixels, one red pixel, and one blue pixel. The emphasis on green emulates the human visual system, which is more sensitive to green wavelengths.

During capture, the camera reads out a row of the image sensor where each pixel voltage is amplified at the column level and then quantized with an ADC. A frame rate determines the time it takes to read and quantize a complete image. The camera emits a digital signal referred to as a RAW image, and sends it to the ISP for processing.

### A.2. Image Signal Processor

Modern mobile devices couple the image sensor with a specialized image signal processor (ISP) chip, which is responsible for transforming the RAW data to a final, compressed image—typically, a JPEG file. ISPs consist of a series of signal processing stages that are designed to make the images more palatable for human vision. While the precise makeup of an ISP pipeline varies, we describe a typical set of stages found in most designs here.

**Denoising.** RAW images suffer from three sources of noise: *shot noise*, due to the physics of light detection; *thermal noise* in the pixels, and *read noise* from the readout circuitry. The ISP uses a denoising algorithm such as BM3D [12] or NLM [4] to improve the image's SNR without blurring important image features such as edges and textures. Denoising algorithms are typically expensive because they utilize spatial context, and it is particularly difficult in low-light scenarios.

**Demosaicing.** The next stage compensates for the image sensor's color filter mosaic. In the Bayer layout, each pixel in the RAW image contains either red, green, or blue data; in the output image, each pixel must contain all three channels. The *demosaicing* algorithm fills in the missing color channels for each pixel by interpolating values from neighboring pixels. Simple interpolation algorithms such as nearest-neighbor or averaging lead to blurry edges and other artifacts, so more advanced demosaicing algorithms use gradient-based information at each pixel to help preserve sharp edge details.

**Color transformations and gamut mapping.** A series of color transformation stages translate the image into a color space that is visually pleasing. These color transformations are local, per-pixel operations given by a 3×3 matrix multiplication. For a given pixel $p \in \mathbb{R}^3$, a linear color transformation is a matrix multiplication:

$$p' = Mp \qquad (1)$$

where $M \in \mathbb{R}^{3\times 3}$.

The first transformations are *color mapping* and *white balancing*. Color mapping reduces the intensity of the green channel to match that of blue and red and includes modifications for artistic effect. The white balancing transformation converts the image's color temperature to match that of the lighting in the scene. The matrix values for these transformations are typically chosen specifically by each camera manufacturer for aesthetic effect.

The next stage is *gamut mapping*, which converts color values captured outside of a display's acceptable color range (but still perceivable to human vision) into acceptable color values. Gamut mapping, unlike the prior stages, is nonlinear (but still per-pixel). ISPs may also transform the image into a non-RGB color space, such as YUV or HSV [41].

**Tone mapping.** The next stage, *tone mapping*, is a nonlinear, per-pixel function with multiple responsibilities. It compresses the image's dynamic range and applies additional aesthetic effects. Typically, this process results in aesthetically pleasing visual contrast for an image, making the dark areas brighter while not overexposing or saturating the bright areas. One type of global tone mapping called **gamma compression** transforms the luminance of a pixel $p$ (in YUV space):

$$p' = Ap^\gamma \qquad (2)$$

where $A > 0$ and $0 < \gamma < 1$. However, most modern ISPs use more computationally expensive, local tone mapping based on contrast or gradient domain methods to enhance image quality, specifically for high dynamic range scenes such as outdoors and bright lighting.

**Compression.** In addition to reducing storage requirements, compression helps reduce the amount of data transmitted between chips. In many systems, all three components—the image sensor, ISP, and application logic—are on physically separate integrated circuits, so communication requires costly off-chip transmission.

The most common image compression standard is JPEG, which uses the discrete cosine transform quantization to exploit signal sparsity in the high-frequency space. Other algorithms, such as JPEG 2000, use the wavelet transform, but the idea is the same: allocate more stage to low-frequency information and omit high-frequency information to sacrifice detail for space efficiency. This JPEG algorithm is typically physically instantiated as a codec that forms a dedicated block of logic on the ISP.

## B. Proposed Pipelines

The main paper describes two potential simplified ISP pipelines including only the stages that are essential for all algorithms we studied: demosaicing, gamma compression, and denoising. Normalized data was shown in the main paper to make more efficient use of space, but here in Figure 8 we show the absolute error for each benchmark. As depicted in the main paper, the pipeline with just demosaicing and gamma compression performs close to the baseline for most benchmarks; the outlier is SGBM, where denoising has a significant effect. OpenFace, also as discussed in the main paper, is alone in performing better on the converted images than on the original dataset.

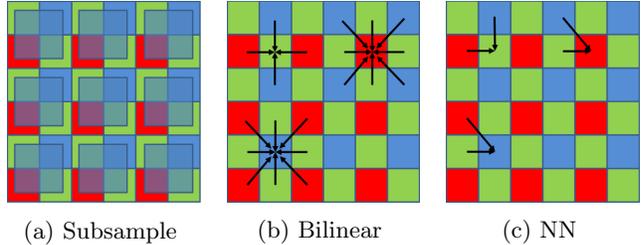

(a) Subsample  (b) Bilinear  (c) NN

Figure 9: Visualizations for the approximate forms of demosaicing: subsampling, bilinear interpolation, and a nearest-neighbor algorithm.

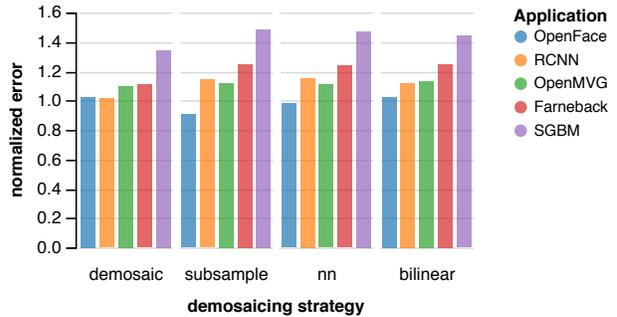

Figure 10: Normalized task error for four demosaicing strategies. Each cluster shows a configuration simulating a pipeline with only gamma compression enabled. The *demosaic* cluster shows the original demosaiced data (i.e., all stages were reversed *except* for demosaicing). The others show images with simulated demosaicing using subsampling (the strategy described in the main paper), nearest-neighbor demosaicing, and bilinear interpolation.

## C. Approximate Demosaicing

We find that the demosaicing ISP stage is useful for the vision applications we examine. In the main paper, we describe *subsampling* as a circuit-level replacement for "true" demosaicing on the ISP.

Here, we also consider two other lower-quality demosaicing techniques that use simpler signal processing. The two techniques are bilinear interpolation and a nearest-neighbor algorithm. Figure 9 visualizes these techniques and Figure 10 shows their results. Our subsample demosaicing fills in missing channel values from their corresponding channels in the Bayer pattern. Bilinear interpolation fills channel values by averaging the corresponding local pixels, and the nearest-neighbor technique simply copies the channel value from a nearby pixel.

Figure 10 compares the vision task performance for these techniques. All three mechanisms lead to similar vision error. For this reason, we chose the cheapest technique, which eliminates the need for any signal processing: subsampling.

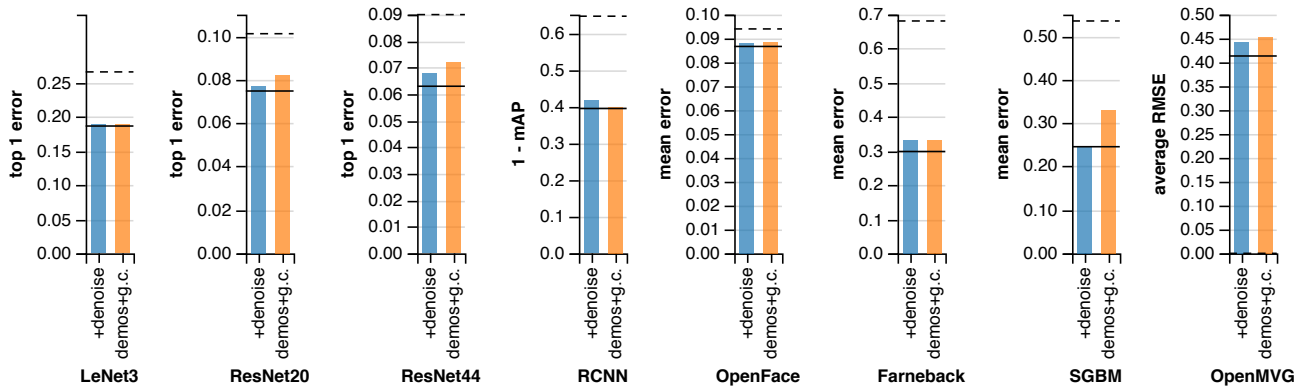

Figure 8: Vision accuracy for two proposed pipelines.

### D. Resolution

While the resolution of a standard mobile system's image sensor can be on the order of a megapixel, the input resolution to a state-of-the-art convolutional neural network is often no more than 300×300. For this reason, images are typically scaled down to fit the input dimensions of the neural network. While the algorithms used to scale down these images are typically edge aware, it is also possible to output a reduced resolution from the image sensor. One method of doing this is pixel-binning which connects multiple photodiodes together, collectively increasing the charge and thereby reducing the error associated with signal amplification [52].

Figure 11 shows the results of our resolution experiments we conducted with the high resolution version of CIFAR-10 dataset that we describe in the main paper. Our testing was conducted by averaging pixels in the region of the sensor which would be binned, thereby reducing resolution. Any further reduction in accuracy was performed with OpenCV's edge aware image scaling algorithm [28]. As can be seen in Figure 11 the increase in error when using pixel binning isn't remarkably large, but we find generally capturing with a higher resultion is always better if possible. This presents a tradeoff between energy used to capture the image and the error for vision tasks.

### E. Quantization

In the main paper, we present logarithmic quantization as a way to replace digital gamma compression in the ISP. As we discussed, the benefit that gamma compression provides is largely to enable a more compressed encoding to represent the intensity values by converting the data distribution from log-normal to normal. However, information theory tells us that we can achieve the minimum quantization error (and maximum entropy) when the encoded distribution is uniform [29]. So, in this section we go further by exploring the possibility of tuning quantization specifically to the statistical properties of natural scenes.

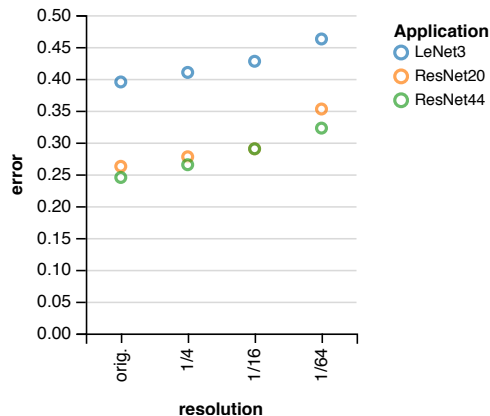

Figure 11: Impact of resolution on three CNNs for object recognition. Using a custom data set consisting of higher-resolution images from ImageNet matching the CIFAR-10 categories, we simulate pixel binning in the sensor, which produces downsampled images. The $y$-axis shows the top-1 error for each network.

To compute the optimal quantization levels for our data, we first fit a log-normal curve to the histogram of natural images. For our experiments we used a subset of CIFAR-10 [32] which had been converted to its raw form using the CRIP tool. This log-normal curve served as our probability density function (PDF), which we then integrated to compute our cumulative density function (CDF). We then inverted the CDF to determine the distribution of quantization levels. Using uniformly distributed values across the CDF results in uniformly distributed encoded (digital) values. Figure 12 shows both the input and quantized distributions.

|  | Demosaic | NL-Means Denoise | Color Transforms | Gamut Map | Tone Map | JPEG Compress |
|---|---|---|---|---|---|---|
| Instructions | $3.45 \times 10^8$ | $4.84 \times 10^{11}$ | $2.40 \times 10^8$ | $5.38 \times 10^8$ | $4.63 \times 10^8$ | $6.74 \times 10^8$ |
| Cycles | $3.62 \times 10^8$ | $3.06 \times 10^{11}$ | $2.26 \times 10^8$ | $8.09 \times 10^8$ | $4.84 \times 10^8$ | $2.94 \times 10^8$ |
| Cache Refs | $4.17 \times 10^6$ | $1.60 \times 10^8$ | $1.80 \times 10^6$ | $4.11 \times 10^6$ | $2.63 \times 10^6$ | $6.96 \times 10^5$ |
| FP Ops | $1.95 \times 10^5$ | $6.77 \times 10^8$ | $1.45 \times 10^5$ | $2.43 \times 10^5$ | $1.52 \times 10^5$ | $9.40 \times 10^3$ |

Table 2: Profiling statistics for software implementations of each ISP pipeline stage.

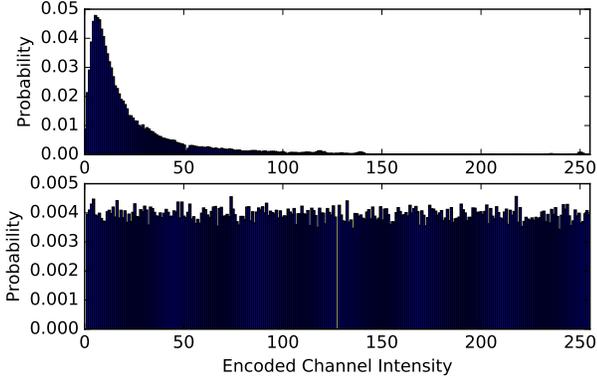

Figure 12: Histograms of the light intensity distribution for CRIP-converted raw CIFAR-10 data (top) and CDF quantized CIFAR-10 data (bottom).

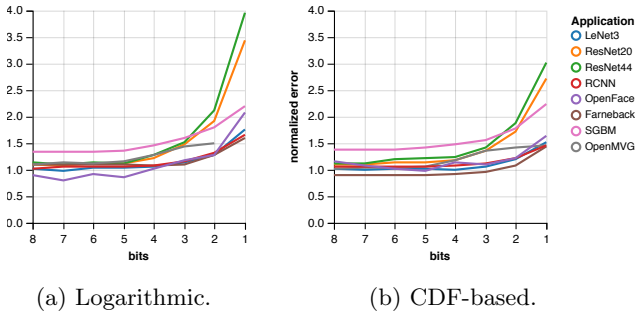

(a) Logarithmic.    (b) CDF-based.

Figure 13: Effect of the image quantization strategy on vision accuracy in a pipeline with only demosaicing enabled. This figure shows logarithmic quantization and a second strategy based on measuring the cumulative distribution function (CDF) of the input data. For traditional linear quantization, see the main paper.

This CDF-based technique approximates the minimum-entropy quantization distribution. An even more precise distribution of levels may be derived using the Lloyd-Max algorithm [20].

Figure 13 compares the vision task performance using this CDF technique with the simpler, data-agnostic logarithmic quantization described in the main paper. The error across all applications tends to be lower. While the logarithmic quantization strategy is less sensitive to bit-width reduction than linear quantization, the CDF technique is even less sensitive. Where 5 bits suffice for most benchmarks under logarithmic quantization, 4 bits generally suffice with CDF-based quantization.

Our main proposal focuses on logarithmic quantization, however, because of hardware feasibility: logarithmic ADCs are known in the literature and can be implemented with a piecewise-linear approximation scheme [30]. Using the CDF quantization scheme would require an ADC with *arbitrary* quantization levels; the hardware complexity for such an ADC design is not clear in the literature.

## F. ISP Profiling

While total energy numbers for ISPs have been published, we are unaware of an energy breakdown per ISP stage. A promising area of future work is to simulate each ISP stage at the hardware level, but as a simpler examination of ISP stage costs, we present measurements based on software profiling. For the description of the experimental setup, see the main paper.

In Table 2, we show the number of dynamic instructions, cycles, L1 cache references, and floating-point operations needed to perform each of the ISP stages. In the table, *instructions* indicates the number of dynamic assembly instructions that the CPU executed, and *cycles* shows the number of CPU cycles elapsed during execution. When the cycle count is smaller than the number of instructions, the CPU has successfully extracted instruction-level parallelism (ILP); otherwise, performance is worsened by lower ILP or frequent memory accesses. The *cache refs* row quantifies the rate of memory accesses: it shows the number of times the L1 cache was accessed, which is an upper bound on the number of access to off-chip main memory. While most operations in a CPU use simple fixed-point and integer arithmetic, complex scientific computation uses floating-point operations, which are more expensive.

The *FP ops* row shows the number of floating-point instructions executed in each stage.

With this level of detail, we see that denoising is a significantly more complex stage than the others. With this one exception, all stages require similar numbers of instructions, cycles, and cache references. The floating-point operation frequency is similar as well, with the exception of the JPEG compression stage: the JPEG codec is optimized for fixed-point implementation. We plan to explore these costs in more detail with a hardware implementation in future work, but these software-based measurements demonstrate that the implementation of the denoising stage will be of particular importance.